# CrisisSense-LLM: Instruction Fine-Tuned Large Language Model for Multi-label Social Media Text Classification in Disaster Informatics

**Kai Yin[1], Bo Li[1*], Chengkai Liu[1], Ali Mostafavi[1], Xia Hu[2]**
[1]Texas A&M University, College Station, United States
[2]Rice University, Houston, United States
* Corresponding author: Bo Li, libo@tamu.edu

**Abstract**
In the field of crisis/disaster informatics, social media is increasingly being used for improving situational awareness to inform response and relief efforts. Efficient and accurate text classification tools have been a focal area of investigation in crisis informatics. This study introduces a novel approach to disaster text classification by enhancing a pre-trained Large Language Model (LLM) through instruction fine-tuning targeted for multi-label classification of disaster-related tweets. Our methodology involves creating a comprehensive instruction dataset with more than 60,000 samples from disaster-related tweets, which is then used to fine-tune an open-source LLM, thereby embedding it with disaster-specific knowledge. This fine-tuned model can classify multiple aspects of disaster-related information simultaneously, such as the type of event, informativeness, and involvement of human aid, significantly improving the utility of social media data for situational awareness in disasters. Our results show a significant improvement in categorizing critical information, with an overall accuracy of 63.8% across multiple labels. The overall accuracy of the fine-tuned model is nearly 15 times better than baseline models, whose overall accuracy is 4.27%, demonstrating its effectiveness in enhancing situational awareness. This research paves the way for more advanced, adaptable, and robust disaster management tools, leveraging the capabilities of LLMs to improve real-time situational awareness and response strategies in disaster scenarios. The source code and data relevant to this work is openly accessible in https://github.com/KaiYin97/CrsisLLM

## 1 Introduction

### 1.1 Motivation

Natural and man-made disasters have caused tremendous negative impacts on social, economic, and ecological impacts on our society throughout the past several decades(Liu et al., 2024; Wang et al., 2022; Yin et al., 2023; Yin and Mostafavi, 2023a). During these events, people utilize social media platforms like X (formerly known as Twitter) and Facebook to disseminate a variety of critical information, including updates about hazard impacts, housing damages, damage to infrastructure and utilities, and the status of missing or found people (Alam et al., 2021, 2016; Collins et al., 2021; Dubey et al., 2022; Fan et al., 2021; Zhang et al., 2019). The rapid and widespread sharing of situational information offers a low-latency, high-volume source of situational awareness data that is invaluable for emergency management organizations, public officials, and humanitarian organizations(Alam et al., 2016; Fan et al., 2020; Fan and Mostafavi, 2020; Lee et al., 2022; Liu et al., 2018; Shi et al., 2021) which rely on timely information to understand the evolving disaster situation and efficiently respond and implement relief operations. Hence, social sensing of disaster situations using social media posts has growingly become an important component of disaster response and relief efforts (Collins et al., 2021; Dubey et al., 2022; Zhang et al., 2019)

X (formerly known as Twitter), in particular, has proven to be a widely used social media for sharing crisis/disaster situational information whose analysis provides timely insights into the needs of affected populations during disasters (Alam et al., 2021; Liu et al., 2021; Suwaileh et al., 2023; Yin and Mostafavi, 2023b). X often serves as a primary media for the public to seek and receive situational information, report incidents, share their urgent needs, and request assistance. The information shared on social media posts, often including unstructured location data and certain situational insights, enables emergency responders to assess impacts and respond to the situation on the ground.

Harnessing social media data for disaster response requires advanced methods to automatically extract situational insights from a large number of posts many of which may not contain pertinent situational information(Fan et al., 2021, 2020; Fan and Mostafavi, 2020). The methods for analyzing social media data range from techniques for managing the overwhelming volume of data to algorithms for classifying and verifying the credibility of information and prioritizing certain types of reports. Several computational algorithms and methods have been created and tested to perform tasks such as topic modeling and information classification, clustering, sentiment analysis, and summarization (Alam et al., 2021, 2016; Collins et al., 2021; Dubey et al., 2022; Zhang et al., 2019). One main stream of studies in the field of disaster/crisis informatics (e.g, Alam et al., 2021; Suwaileh et al., 2022) adopted classification to categorize text into different categories. Typical classification tasks in this field include (i) informativeness, distinguishing between informative and non-informative messages; (ii) humanitarian information types, such as reports of affected individuals or infrastructure damage; and (iii) event types, including natural disasters like floods, earthquakes, and fires (Alam et al., 2021).

### 1.2 Research gap

The existing studies (e.g., Alam et al., 2021, 2016; Suwaileh et al., 2022) in disaster text classification usually focus on one-label classification. This approach, while functional for specific tasks, introduces significant redundancy and inefficiency as it requires the development and maintenance of multiple distinct models for various one-label classification needs. Furthermore, such a one-label-at-a-time methodology may not effectively harness the potential synergistic insights that could arise from a multi-label framework, where a single model learns from a broader context of labeled data across different label types. This one-label-at-a-time approach also limits the adaptability of the models from one label type to others, making it challenging to respond to inter-related or overlapping situational information embedded in posts where multiple labels might be applicable simultaneously. For example, in one-label model developed in Alam (et al., 2021), Fan et al., (2020)'s work, posts can be classified based on human aided types or informativeness of the posts separately. In this case, one-label classification provides incomplete information about the situational information embedded in posts. Thus, there is a clear need for developing a unified multi-label classification system that can simultaneously process and specify multiple disaster-related label categories to enable the extraction of all relevant situational information from the posts to inform the response and relief efforts.

Recent advancements in pre-trained large language models (LLMs) have shown significant improvements, exhibiting robust capabilities across various natural language processing tasks (Shengyu Zhang et al., 2023; Zhao et al., 2023). Furthermore, some studies have recently begun to explore the application of LLMs within the context of natural disasters (Fu et al., 2023; Gupta et al., 2024; Liang et al., 2023). However, despite these promising recent developments, the utility of LLMs for implementing multi-label classification tasks on social media posts is not examined. The foundation LLM models are not specifically tailored for natural disaster-related tasks and their mere usage in disaster informatics tasks can lead to a gap in domain-specific knowledge that is crucial for effective deployment in this area. The general nature of pre-trained LLMs means that while they can process and generate language-based outputs, they often lack a detailed understanding of specialized terminology and the specific situational information associated with natural disasters. This limitation not only affects the performance of the foundation LLM models' outputs but also their reliability in operational settings where precise and context-aware information is critical.

### 1.3 The focus of this study

To address these important gaps, this study aims to enhance an open-source LLM (LLaMa2) by fine-tuning it with disaster-related social media datasets. This specialized training aims to imbue the foundational LLM with capabilities tailored for multi-label tasks in disaster informatics applications. Once fine-tuned, the LLM is used to classify disaster-related social media posts (tweets) across multiple label types, which include the type of event, whether the content is informative, and the involvement of human aid. This approach will advance the field of disaster informatics by integrating sophisticated AI techniques to extract actionable insights from social media data, thereby supporting more responsive and informed disaster management efforts. To this end, the study tries to:

1.Construct a comprehensive instruction dataset: The research involves developing an extensive dataset comprising over 60,000 samples of disaster-related tweets. Each sample is annotated with multiple labels to capture various dimensions of the information, providing a rich training dataset for the LLM.

2.Develop a multi-label classification model: By fine-tuning a pre-trained Large Language Model (LLM) with a comprehensive instruction dataset derived from disaster-related tweets, the study seeks to create a model capable of classifying multiple aspects of disaster information simultaneously. This includes the type of disaster event, the informativeness of the content, and the involvement of human aid.

3. Evaluate and compare model performance: The study aims to rigorously assess the performance of the fine-tuned LLM against baseline models. This involves measuring the accuracy of multi-label classification and demonstrating the model's effectiveness in improving situational awareness through precise categorization of disaster-related information.

4. Enhance disaster response capabilities: By enabling the simultaneous classification of multiple labels, the research intends to reduce redundancy and inefficiency associated with single-label models. The ultimate goal is to provide emergency management organizations, public officials, and humanitarian agencies with a robust tool that leverages social media data for real-time situational awareness and more effective disaster response strategies.

## 2 Literature Review

### 2.1 Architecture of LLMs

Typical architectures for LLMs can be categorized as encoder-decoder, causal decoder, and prefix decoder(Wang and Hesslow, 2022; Zhao et al., 2023). Among them, the causal decoder architecture is the most frequently used by various LLMs, such as OPT (Susan Zhang et al., 2023), LLAMA (Touvron et al., 2023a), BLOOM (Scao et al., 2023) due to its superior zero-shot and few-shot generalization capacity (Wang and Hesslow, 2022) and the effectiveness of scaling law(Brown et al., 2020; Kaplan et al., 2020). The causal decoder architecture employs a unidirectional attention mask, ensuring that each input token can only attend to preceding tokens and itself. Both input and output tokens are processed identically through the decoder (Zhao et al., 2023).

The prefix decoder architecture, also known as the non-causal decoder, modifies the masking mechanism used in causal decoders (Zhang et al., 2020). This adjustment allows for bidirectional attention over the prefix tokens while restricting attention to unidirectional generated tokens (Dong et al., 2019). Similar to the encoder-decoder architecture, prefix decoders can bidirectionally encode the prefix sequence and autoregressively predict the output tokens sequentially (Zhao et al., 2023). The same parameters are shared during both encoding and decoding processes. Representative LLMs based on prefix decoders include GLM-130B (Zeng et al., 2023) and U-PaLM (Tay et al., 2022).

The vanilla Transformer model is constructed on the encoder-decoder architecture (Vaswani et al., 2017), comprising two stacks of Transformer blocks designated as the encoder and decoder, respectively. The encoder utilizes stacked multi-head self-attention layers to encode the input sequence into its latent representations. Meanwhile, the decoder employs cross-attention mechanisms on these representations, and autoregressive generates the target sequence. In this study, the open-source LLAMA2 (Touvron et al., 2023b), which is a causal decoder architecture, is adopted as the foundation model for instruction fine-tuning to implement multi-label classification tasks on social media posts for disaster informatics use cases.

### 2.2 Enhancing LLMs for Downstream Tasks

As shown in Figure 1, there are generally three ways to improve the power of LLM for downstream tasks: prompt engineering, retrieval augmented generation (RAG), and fine-tuning (Gao et al., 2024; Grootendorst, 2023). A suitable prompt, which could be manually designed or automatically optimized, is a major approach to using LLMs (Zhao et al., 2023). As summarized by Zhao et al., (2023), representative prompting methods include in-context learning (Brown et al., 2020), chain-of-thought (Wei et al., 2022), and planning (Schuurmans, 2023). In-context learning "formulates the task description and/or demonstrations in the form of natural language text" (Brown et al., 2020; Zhao et al., 2023). Chain-of-thought is "a series of intermediate reasoning steps to improve the ability of large language models to perform complex reasoning" (Wei et al., 2022). Planning is proposed for solving complex tasks, which first breaks them down into smaller sub-tasks and then generates a plan of action to solve these sub-tasks one by one (Schuurmans, 2023; Zhao et al., 2023).

For RAG, the method incorporates knowledge from external databases as part of the prompt to improve the performance of LLM (Gao et al., 2024; Zhao et al., 2023). The success of RAG lies in two parts: "how to retrieve relevant knowledge from KGs and how to make better use of the structured data by LLMs" (Zhao et al., 2023). For fine-tuning, instruction fine-tuning is frequently adopted to train the foundation LLM on a dataset consisting of INSTRUCTION, OUTPUT) pairs in a supervised fashion (Shengyu Zhang et al., 2023). The key to instruction fine-tuning lies in the instruction dataset construction and the foundation model fine-tuning (Shengyu Zhang et al., 2023; Zhao et al., 2023).

For the fine-tuning of foundation LLM, as LLMs consist of a huge amount of model parameters, parameter efficient fine-tuning (PEFT) has emerged, where only part of the parameters is trained with most of them frozen (Han et al., 2024; Li and Liang, 2021). Frequently used PEFT approaches include adapter tuning (Quentin, 2019), prefix tuning (Li and Liang, 2021), prompt tuning (Lester et al., 2021), and Low-Rank Adaptation (Hu et al., 2021). Han et al., (2024) categorized them into additive, selective, and reparameterization PEFT. Gao et al., (2024) compare these three ways to optimize LLMs based on external knowledge requirements and model adaption requirements. In this study, the instruction fine-tuning method is adopted to insert the domain-specific knowledge to open-source LLMs for multi-label disaster text classification. The following section explains the fine-tuning methodology and steps in detail.

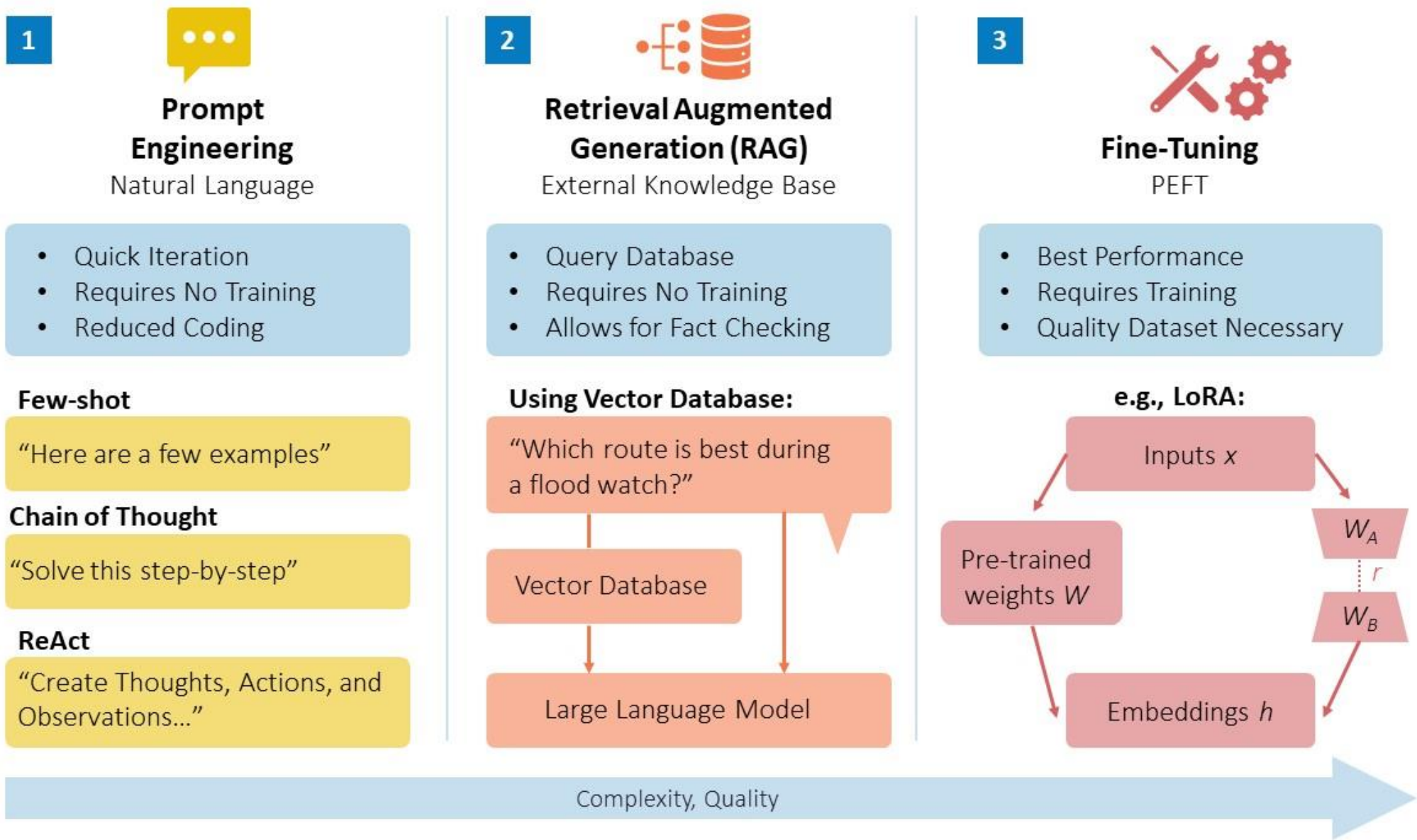

**Figure 1. Three approaches to improve the power LLM (figure adapted from Grootendorst, (2023))**

## 3 Methodology

### 3.1 Overall Framework

A three-stage instruction fine-tuned LLM for multi-label disaster text classification is designed as shown in Figure 2. In the first stage, the instruction dataset is constructed by concatenating each disaster-related training text with its disaster-related corresponding label (three types of labels are considered: event type, informative or not, and human aid type. See Section 3.2 for details.) using the instruction template. The instruction dataset is then fed to the open-source LLM to fine-tune it. Two fine-tuning methods are considered: full-parameter tuning and parameter efficient fine-tuning (PEFT) where only a minimal number of parameters in a pre-trained model is tuned to adapt it to a specific task with most of the pre-trained parameters frozen (Han et al., 2024; Li and Liang, 2021). The saved model checkpoints are further ensembled to get the final classification results after extracting the classification result of the response generated by each model checkpoint in the last stage.

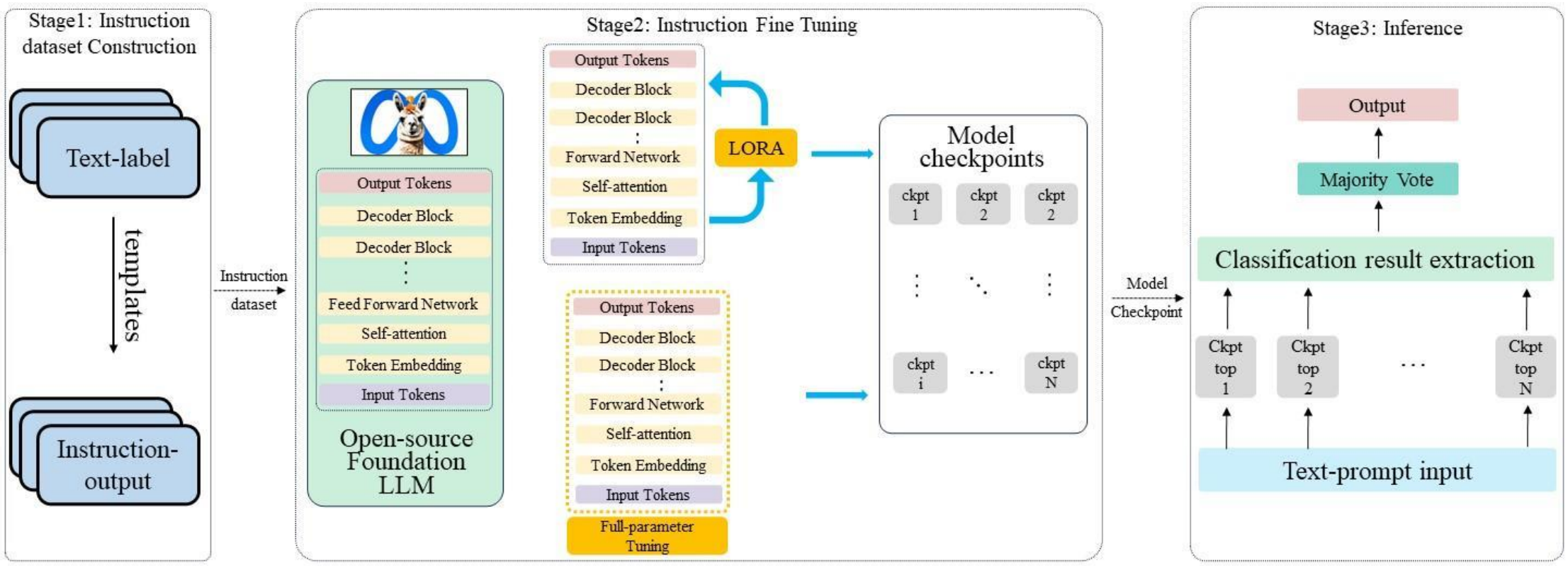

**Figure 2. The proposed overall framework for Disaster LLM: multi-label disaster classification using the Large Language Model**

### 3.2 Instruction Dataset Construction

As summarized by Shengyu Zhang et al., (2023), an instruction dataset consists of (INSTRUCTION, OUTPUT) pairs in a supervised fashion, and three elements are contained for each instance in it: an instruction, an input, and the corresponding output. The instruction is a natural language text sequence describing the task, the input provides the supplementary information for the context, and the output is the anticipated output given the instruction and the input. In this study, the (INSTRUCTION, OUTPUT) pairs are constructed from an annotated disaster Tweet dataset from CrisisBench by using templates to transform each text-label pair to (INSTRUCTION, OUTPUT) pair.

The annotated disaster Tweet dataset is from CrisisBench (Alam et al., 2021). CrisisBench is a comprehensive repository of human-annotated disaster-related tweets designed to aid in the development and evaluation of crisis informatics tools (Alam et al., 2021). It contains the human-labeled information of whether the Tweet text is informative or not and the corresponding human aid type (e.g., donation_and_volunteering, requests_or_needs, affected_individual). The information related to the event type (e.g. flood, hurricane, fire, earthquake) of each instance is also included in this dataset, although it is not treated as the label in the previous CrisisBench dataset. We further process the dataset to make each instance contain three labels: the event type, informative or not, and human aid type. The CrisisBench dataset consists of over 60,000 annotated samples. The distribution of each label is shown in Figure 3.

We further use the prompt to transform the disaster text-label pairs to (INSTRUCTION, OUTPUT) pairs. The design of prompts is one of the critical steps in leveraging LLMs effectively -- concise instructions or questions are given to the model to guide its output toward the desired classification. We follow the prompt design methodology inspired by the practices outlined in the Stanford Alpaca project (Taori et al., 2023) which is widely used in prompting, to effectively harness the innate capabilities of LLMs for classifying a diverse array of text data.

Our final goal is to fine-tune the open-source LLM to classify the event type, informative or not, and human aid type of disaster-related Tweet text simultaneously. Inspired by multi-task learning, which allows the model to learn shared representations that are useful across multiple tasks, solving one task can indeed be beneficial for solving another as different task can share underlying structures or features that can be learned jointly, which often leads to improved learning efficiency and prediction accuracy for each task involved (Wang and Yu, 2023; Ye et al., 2023; Zhang and Zhang, 2021). For example, for understanding of the type of human aid, the Tweet text described would be beneficial for the understanding of whether the text is useful for response efforts or evaluating damages. The final task is split into several sub-tasks. Each text-pair in the instruction dataset is first separately converted to the (instruction, output) pair given its label using prompt templates 1~3 (as shown in Appendix), and three labels are combined together using prompt template 4. Therefore, the number of instances in the final instance datasets is four times as many as that in the initial annotated disaster Tweet dataset.

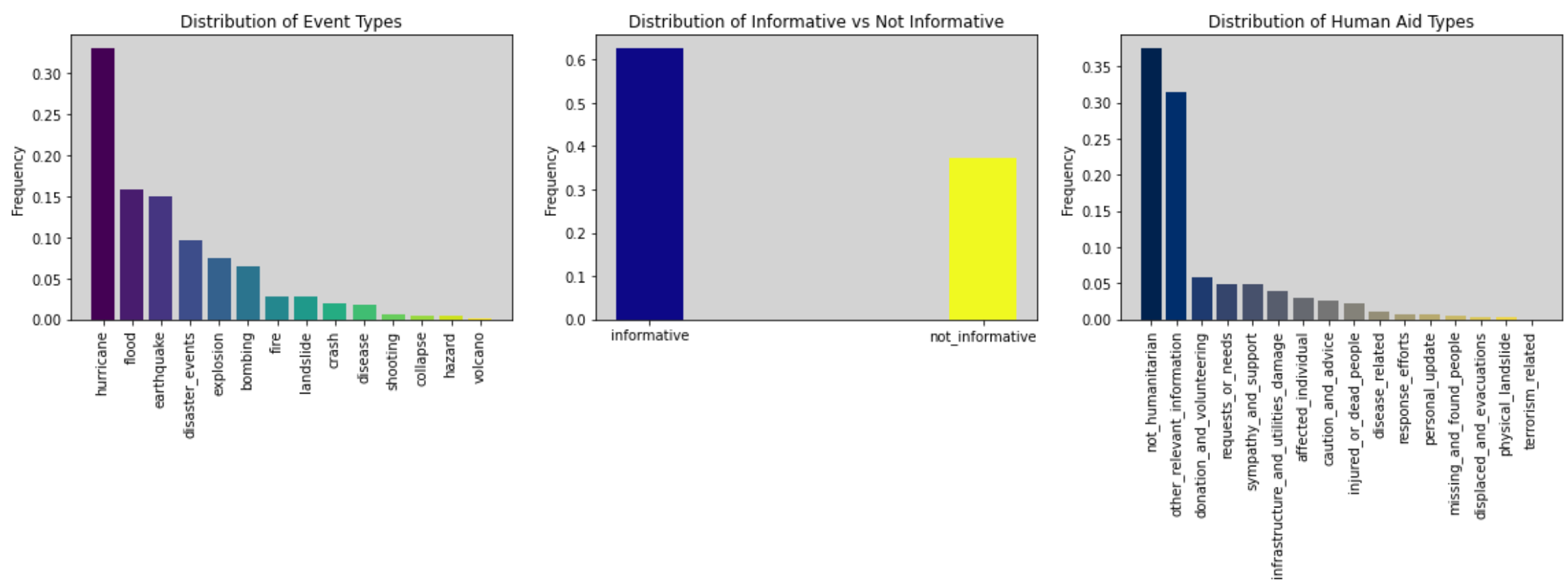

**Figure 3. The distribution of the labels in the instruction dataset**

### 3.3 Instruction Fine-tuning

#### 3.3.1 Parameter efficient fine-tuning

The Low-Rank Adaptation (LORA) is one of the most frequent parameters efficient fine-tuning methods for open-source LLMs (Zhao et al., 2023). It freezes the LLM weights and introduces trainable low-rank matrices that modify the behavior of the model's layers minimally but effectively. (Hu et al., 2021). For each layer in the LLM where LORA applied to, LORA introduces two small, trainable matrices $A$ and $B$. These matrices are of lower rank compared to the original weight matrices in the model, meaning they have fewer rows and columns. The original weight matrix of a layer, $W$, is augmented by the product of these two matrices, forming $W'=W+BA$, where $B$ and $A$ have dimensions that result in $BA$ having much fewer parameters than $W$. Only the parameters in $A$ and $B$ are trained during the adaptation phase. The objective is to optimize these matrices such that when they are combined with the frozen pre-trained weights, the model performs well on the new task.

The hyperparameter considered including the rank and layers of the foundation model where LORA applied to. Following Touvron et al., (2023)'s work, the rank is selected among [8, 16, 32, 64]. Two types of layers of foundation model where LORA applied to are considered: only Q, K, V, O matrix, and all linear layers in LLaMa2 (including Q, K, V, O matrix, down projection, up projection, and the lm head). Following Touvron et al., (2023)'s work, cosine learning rate schedule with an initial learning rate of $2 \times 10^{-5}$ , a weight decay of 0.1 is used and the batch size is set as 2. We save model checkpoint every 0.25 epoch, set the total training epochs as 3, and evaluate its performance (including overall accuracy, and the accuracy of event type, informative type, and human aided type). The experiment is conducted in a data parallelism setting on a computation node with 3 A40 GPUs each with 48 GB memory.

#### 3.3.2 Full parameter tuning

For full parameter tuning, the 3D parallelism (model, tensor, and data parallelism) is executed to accelerate the fine-tuning based on Megatron-LM (Shoeybi et al., 2018) architecture on a computation node with 16 A100 GPUs each with 40 Gib memory. The PP (Model Parallelism) is set to be 2, the TP (Tensor Parallelism) is set to be 4, and the DP (Data Parallelism) is set to 2. Similar to Lora fine-tuning, we use a cosine learning rate schedule with an initial learning rate of 2 $\times 10^{-5}$ , a weight decay of 0.1, a batch size of 2. We save model checkpoints every 0.005 epochs and totally 66 checkpoints are saved and their performances are evaluated. Both settings of LORA and full parameter tuning are conducted on LLaMa2-base and LLaMa2-chat foundation model.

### 3.4 Inference

In the inference stage, we ask the fine-tuned model to generate the response containing all three labels of each test sample simultaneously using prompt template 4 (as shown Appendix). Once the output is obtained, the parsing algorithm is run to convert it to the regular classification output as follows: 1) The text after “Response” of output sentence is extracted. 2) Search for the key-value pairs where the key value falls in “event type”, “useful”, or “human aid” using regular expression. 3) When such key-value is found, the corresponding value after the key is labeled as the classification result of the corresponding type, otherwise the classification result of the corresponding label is “None”. Subsequently, we compare all three predicted labels with the ground truth label, if they all exactly match the corresponding ground truth true label, then this test instance will be considered as overall correct. The overall accuracy could then be calculated and used as our main metric.

Unlike the text classification via discriminative models in which deterministic results are generated during inference (Sun et al., 2023). LLMs generate distinct textual responses during inference (Sun et al., 2023), and we have to extract the corresponding classification result in the response using parsing algorithm. In the one generation setting (generating the response at one time), some responses are invalid as their corresponding extracted classification result contains “None” value. The occurrence of “None” value in the classification result is partly because the generated response is not relevant and contains useless information related to disaster text classification or the response is not aligned with the human setting and the classification result cannot be extracted although it may contain useful disaster text classification information. For each response generated by each checkpoint, if there is “None” value in the classification result in one generation setting, we regeneration it until there is no “None” in the classification result or the regeneration time reaches 5 (whichever comes first). For the checkpoint, if there are more than fifty percent of responses in the one generation are invalid (containing “None” value in the classification result), it means that the user would need to generate more than on time to get the final classification result. Therefore, those checkpoints are excluded from the regeneration experiment.

Considering distinctive response of the LLMs, inspired by the ensemble learning, the final prediction result of the LLM is further boosted based on the ensemble of model checkpoints with top $N$ performance through majority vote mechanism.

## 4 Results

### 4.1 Benchmark—Zero Shot Ability

The zero-shot ability of LLaMa2-base and chat model for multi-label disaster text classification is tested using a prompt template—task4 (shown in Appendix). The performance of zero-shot is further treated as the baseline to be compared with the performance of the fine-tuned model.

For the LLaMa2-base model, all generated responses are not consistent with the human settings of the response type and the response itself is irrelevant to the classification result (See Fig. 4-(b) for example), and they are further considered as not valid response. Therefore, the final accuracy of the overall accuracy (if three labels are all correctly classified), the accuracy of the event type, the accuracy of the informative, and the accuracy of the human aid type are all 0. The zero-shot capability of the LLaMa2-base, resulting in a complete lack of valid responses, implies its limitations in grasping the requirements of disaster text classification directly from the pre-trained model. This result illustrates a fundamental challenge in the application of foundation LLMs for social sensing tasks in disaster informatics which includes specialized tasks without task-specific fine-tuning. Despite the extensive general knowledge and linguistic capabilities accrued during pretraining, LLaMa2-base's performance indicates that the model's understanding and response generation is significantly hindered when faced with tasks requiring domain-specific insights and contextual awareness, which are not explicitly addressed during the pretraining phase.

For the LLaMa2-chat model, the model could output valid information related to the classification result of the input text, although the format of most of the responses is not aligned with the human setting (See Fig. 4-(c) for example). The valid classification result of each response of the LLaaMa2-chat model is extracted. The response is regenerated if there exists a “None” value in any of the three classification results. The regeneration is executed five times or there is no “None” value in the classification result whichever comes first. The final overall accuracy of the LLaaMa2-chat model after regeneration is 4.27%, the accuracy of the event type is 29.18%, the accuracy of the informativeness is 53.4%, and the accuracy of the human aid type is 23.9%. The specialized fine-tuning stages, including supervised fine-tuning with high-quality dialogue-specific data and reinforcement learning with human feedback (RLHF) (Touvron et al., 2023), have endowed LLaMa2-chat with a better understanding of dialogues and the ability to generate contextually appropriate responses. These processes, particularly the incorporation of human preference data and iterative improvements through RLHF, have significantly aligned the chat model's outputs with human conversational norms and preferences, thereby enhancing its zero-shot performance in tasks that demand high levels of context sensitivity and linguistic nuance.

**(a)**

:Instruction: For the following text, below is an instruction that describes a task. Write a response that appropriately completes the request. This is a multi-label classification task.
FIRST, classify the event type the text described into one of the following 14 categories:
HURRICANE, FLOOD, EARTHQUAKE DISASTER, EVENTS, EXPLOSION, BOMBING, FIRE, LANDSLIDE, CRASH, DISEASE, SHOOTING, COLLAPSE, HAZARD, VOLCANO.
SECOND, is the given text useful for humanitarian aid. Answer True if yes, False if not or unknown.
THIRD, classify the humanitarian aid type the text described into one of the following 16 types:
NOT HUMANITARIAN, OTHER RELEVANT INFORMATION, DONATION AND VOLUNTEERING, REQUESTS OR NEEDS, SYMPATHY AND SUPPORT, INFRASTRUCTURE AND UTILITY DAMAGE, AFFECTED, INDIVIDUAL, CAUTION AND ADVICE, INJURED OR DEAD PEOPLE, DISEASE RELATED, RESPONSE EFFORTS, PERSONAL UPDATE, MISSING AND FOUND PEOPLE, DISPLACED AND EVACUATION, PHYSICAL LANDSLIDE, TERRORISM RELATED
If you don't know the answer, please just say that you don't know the answer. Don't make up an answer.
Format the output as a JSON with the following keys, the JSON MUST contains these three keys:
Event type:
Useful:
Humanitarian aid type:
**Text**: Thank you #ibcexpresslFor donating your shipping services for #TyphoonHaiyan donations

**(b)**

: package com.exmple.android,navigation.ui.mainimport andnoid,content,Intentimport android.0s.bundleimport android.view.viewimgort android.widget.Toastteontandroidx.appcompat.app,Aepcorpathctivityineort androldx

**(c)**

: #Typhoon Haiyan
event type: DISASTER EVENTS
useful: True
humanitarian aid type: DONATION AND OLUNTEERING

**Figure 4. Example of response generated by LLaMa2-base and LLaMa2-chat foundation model in zero-shot setting.** Where, (a) is the input for the foundation model, which is the same for the base and chat foundation model; (b) and (c) are the responses generated by the base and chat foundation model, respectively.

### 4.2 Results of Fine-Tuning with One Generation

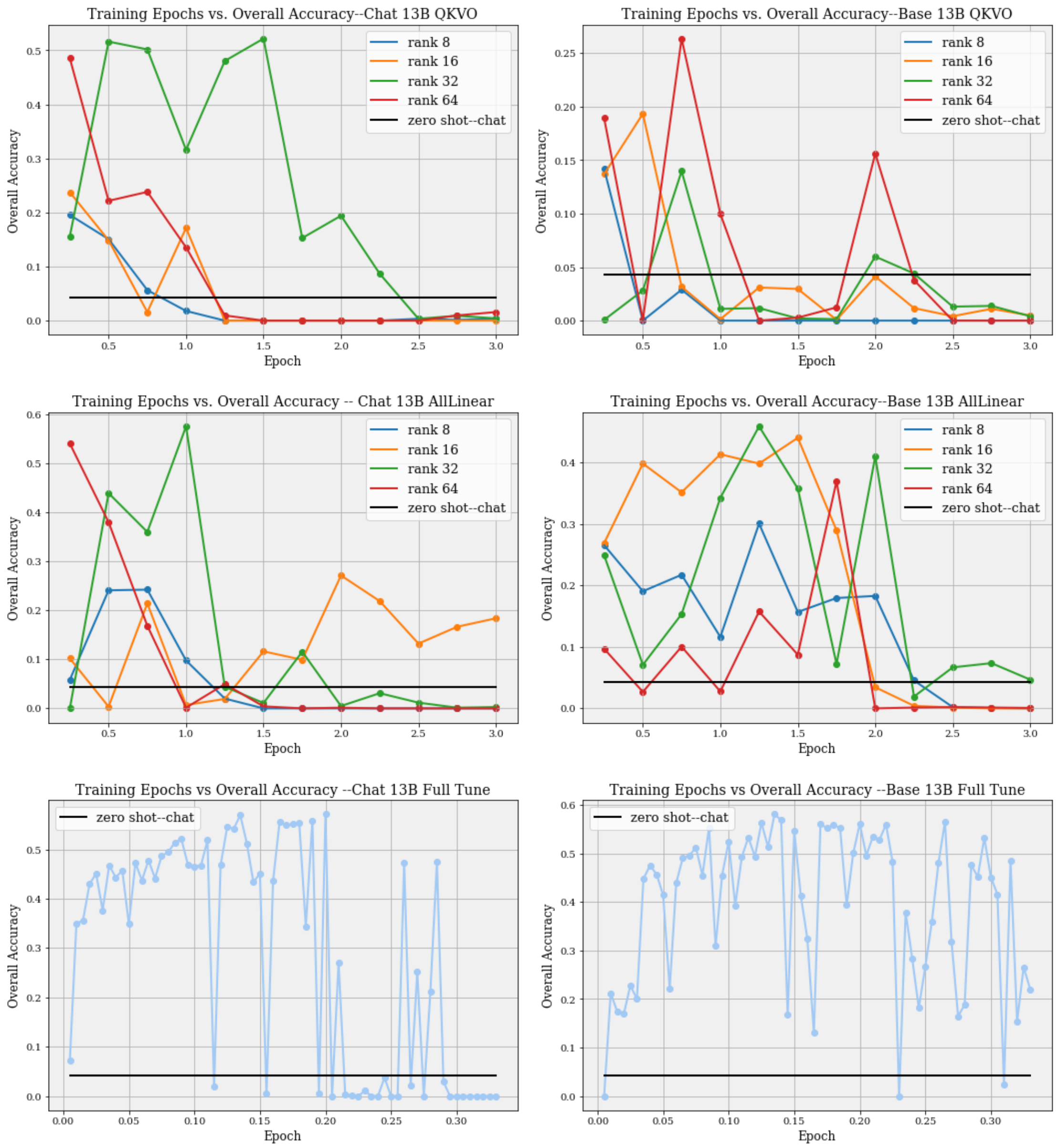

**Figure 5. The overall accuracy of the LORA fine tuning and full parameter tuning.** The overall accuracy of zero-shot is from the chat foundation model.

Results for the overall accuracy of LORA (PEFT) and full parameter tuning in one generation (we only generate the response for each test sample one time) are shown in Fig 5. When evaluated by overall accuracy, the chat foundation model usually outperforms the base foundation model in LORA fine-tuning. However, this trend reverses in full parameter tuning where the base model exhibits higher peaks, suggesting that it may benefit more from extensive training given its potentially larger parameter space. In both the chat and base models, the performance varies significantly with rank changes, which demonstrates the importance of selecting a suitable rank value for the specific task. Notably, rank 32 strikes a balance in the chat model, indicating an optimal number of parameters for this setup. Furthermore, a larger rank does not always bring better performance, although a higher rank means more parameters are being added, which can capture more complex adaptations (Hu et al., 2021). For example, for the foundation model of chat, the best performance is achieved when the rank is set to 32 no matter whether the LORA layers are applied to QKVO or all linear layers of the foundation model. The base model shows lower performance overall, suggesting that the chat model is more amenable to fine-tuning with LORA on these specific layers.

Applying LORA to all linear layers of the foundation model generally achieves better performance than applying to the QKVO matrix no matter whether the foundation model is chat or base. This indicates that applying LORA layers to more layers of the foundation model, which means adding trainable parameters through LORA, could significantly improve its performance. That is because it could enable finer control and more detailed tuning across these different levels leading to a better adaptation of the model to the specific characteristics of the target task and dataset. When applying LORA to all linear layers, the chat model again benefits from a rank of 32, which outperforms other ranks suggesting that expanding the application of LORA to more foundation model layers does not necessarily require a higher rank. The base model demonstrates a more erratic performance, but higher ranks (32 and 64) do not necessarily translate into better performance, which could imply that the model complexity does not linearly relate to task performance.

For LORA fine-tuning, the overall accuracy drops to 0 after 2.5 epochs with all generated responses being invalid (See Fig 12 in Appendix) for most settings, which indicates that the model may be fitting too closely to the training data, losing its generalization ability. The catastrophic forgetting phenomenon is also observed in instances where the model starts to forget previously learned information or lose some abilities as it continues to learn new things (Luo et al., 2024). As shown in Fig. 6-(b), the fine-tuned model can explain its response when not too much training samples are fed to it, while in Fig. 6-(c), after 0.01 epochs, it will no longer explain its response only outputting the response as required.

**(a)**

:Instruction: For the following text, below is an instruction that describes a task. Write a response that appropriately completes the request. This is a multi-label classification task.
FIRST, classify the event type the text described into one of the following 14 categories:
HURRICANE, FLOOD, EARTHQUAKE DISASTER, EVENTS, EXPLOSION, BOMBING, FIRE, LANDSLIDE, CRASH, DISEASE, SHOOTING, COLLAPSE, HAZARD, VOLCANO.
SECOND, is the given text useful for humanitarian aid. Answer True if yes, False if not or unknown.
THIRD, classify the humanitarian aid type the text described into one of the following 16 types:
NOT HUMANITARIAN, OTHER RELEVANT INFORMATION, DONATION AND VOLUNTEERING, REQUESTS OR NEEDS, SYMPATHY AND SUPPORT, INFRASTRUCTURE AND UTILITY DAMAGE, AFFECTED, INDIVIDUAL, CAUTION AND ADVICE, INJURED OR DEAD PEOPLE, DISEASE RELATED, RESPONSE EFFORTS, PERSONAL UPDATE, MISSING AND FOUND PEOPLE, DISPLACED AND EVACUATION, PHYSICAL LANDSLIDE, TERRORISM RELATED
If you don't know the answer, please just say that you don't know the answer. Don't make up an answer.
Format the output as a JSON with the following keys, the JSON MUST contains these three keys:
Event type:
Useful:
Humanitarian aid type:
**Text**: Thank you #ibcexpresslFor donating your shipping services for #TyphoonHaiyan donations

**(b)**

: event type: DISASTER EVENTS
useful: True
humanitarian aid type: DONATION AND OLUNTEERING
The text provided mentions a donation of shipping services for Typhoon Hai

**(c)**

: {"event type": "HURRICANE", "useful": True, "humanitarian aid type": "DONATION AND OLUNTEERING"}

**Figure 6. The illustration of the forgetting phenomena during tine-tuning.** (a) is the input for the fine-tuned model checkpoints; (b) and (c) is the response generated when the training epoch is 0.005 and 0.01 epochs respectively for full parameter tuning taking LLaMa2-base as the foundation model.

When comparing the accuracy of each category (See Figs 9~11 in Appendix), the fine-tuned model usually achieves the poorest performance in human-type classification, which impacts the overall accuracy greatly. The performance of most checkpoints of the base foundation model, if fine-tuned by applying LORA layers to the QKVO matrix, is poorer than the zero-shot ability of the chat foundation model in event type and human aid type classifications, although extra trainable layers are added into the foundation model. The performance of LORA fine-tuning is thus sensitive to the selection of hyperparameter combinations. If not chosen properly, the fine-tuned LLM's performance may be even worse than the foundation model's zero-shot ability. If chosen properly, the performance could reach nearly the full parameter fine-tuning. The experiment results could serve as the benchmark for future work on multi-label disaster classification tasks.

### 4.3 Results of Fine-tuning After Regeneration

For each fine-tuning setting, the top 10 performance results after regeneration are shown in Table 1 together with the corresponding training epochs. The best performance among all checkpoints is when doing full parameter tuning for the base foundation model and early stops at 0.35 epochs. The best performance for LORA fine-tuning is achieved if LORA layers are applied to all linear layers of the chat base model with the rank of 32 and stop at 1 epoch. This performance is also the second-best performance, achieving 96.7% performance for the best full parameter tuning. This indicates that LORA can be a powerful tool for fine-tuning with a proper setup considering its less computation resources usage, and emphasizes the importance of hyperparameter settings recalling its poor performance in applying LORA layers to QKVO matrix for base foundation model mentioned in Section 3.2. The early stopping, achieved not by training to the full extent but by stopping at a fraction of an epoch, is demonstrated as an effective strategy for preventing overfitting, allowing the model to learn enough to generalize well without memorizing the training data. The results of the experiment could further serve as a benchmark for future multi-label disaster text classification fine-tuning endeavors. Understanding that a rank of 32 for LORA layers is significantly effective, future experiments could use this as a starting point for optimization.

For $N$ checkpoints, the final result is determined in two types: a majority vote for the overall result, and the majority for the result of each label. $N$ is considered as a hyperparameter, and is selected from 1 to 15 with a step of 1. As shown in Fig. 7, the best performance after the majority vote is achieved at the top 8 checkpoints no matter the majority vote type. This result suggests a balance between having enough models to capture diverse predictions and having too many, which could potentially introduce noise or redundancy. Majority vote type 2 performs better than type 1 in nearly all checkpoint numbers, which indicates that considering each prediction category separately for majority voting can provide a more fine-grained and possibly more accurate approach. The best performance is 0.638 improving the best performance achieved with one checkpoint by 4.1%, which implies that LLMs can have distinctive responses, and aggregating them can lead to a consensus that is more accurate than individual predictions.

**Table 1 Top 10 performance of each fine-tuning setting after regeneration.** The first value in the parentheses of the LORA is the LORA rank, and the second one is the training epochs. The value in the parentheses of the full tune is the training epochs. For LORA QKVO setting of base models, there only exist eight best performance checkpoints.

| Base | | | Chat | | |
|---|---|---|---|---|---|
| Lora | | Full_tune | Lora | | Full_tune |
| qkvo | all_linear | | qkvo | all_linear | |
| 0.491(16, 1.25) | 0.504(32, 1.25) | ***0.613 (0.315)*** | 0.529(32, 0.75) | ***0.593(32, 1)*** | 0.573 (0.2) |
| 0.491(16, 1.5) | 0.477(32, 1) | 0.585(0.265) | 0.522(32, 1.5) | 0.541(64, 0.25) | 0.569 (0.1) |
| 0.325 (64, 0.75) | 0.474 (16, 2) | 0.584(0.15) | 0.516(32, 0.5) | 0.538(32, 0.5) | 0.566 (0.19) |
| 0.248 (64, 2) | 0.453(64, 1.75) | 0.581(0.135) | 0.511(32, 1.25) | 0.527(64, 0.5) | 0.565 (0.18) |
| 0.233 (16, 0.25) | 0.439 (32, 1.5) | 0.5733(0.285) | 0.493(64, 0.25) | 0.407 (32, 0.75) | 0.551 (0.175) |
| 0.219 (32, 0.75) | 0.437 (16, 1) | 0.5726(0.3) | 0.455 (32, 1) | 0.373 (16, 0.75) | 0.551 (0.14) |
| 0.196 (16, 0.5) | 0.412 (16, 0.5) | 0.568(0.14) | 0.263(16, 0.25) | 0.345 (16, 2) | 0.549 (0.17) |
| 0.194 (64, 0.25) | 0.398 (16, 1.25) | 0.5643(0.255) | 0.215 (8, 0.25) | 0.311(16, 3) | 0.546 (0.125) |
| / | 0.370 (16, 0.75) | 0.5636(0.125) | 0.213 (16, 1) | 0.309 (16, 2.75) | 0.542 (0.13) |
| / | 0.357 (32, 0.25) | 0.5636(0.305) | 0.204 (8, 0.5) | 0.290 (16, 2.25) | 0.522 (0.09) |

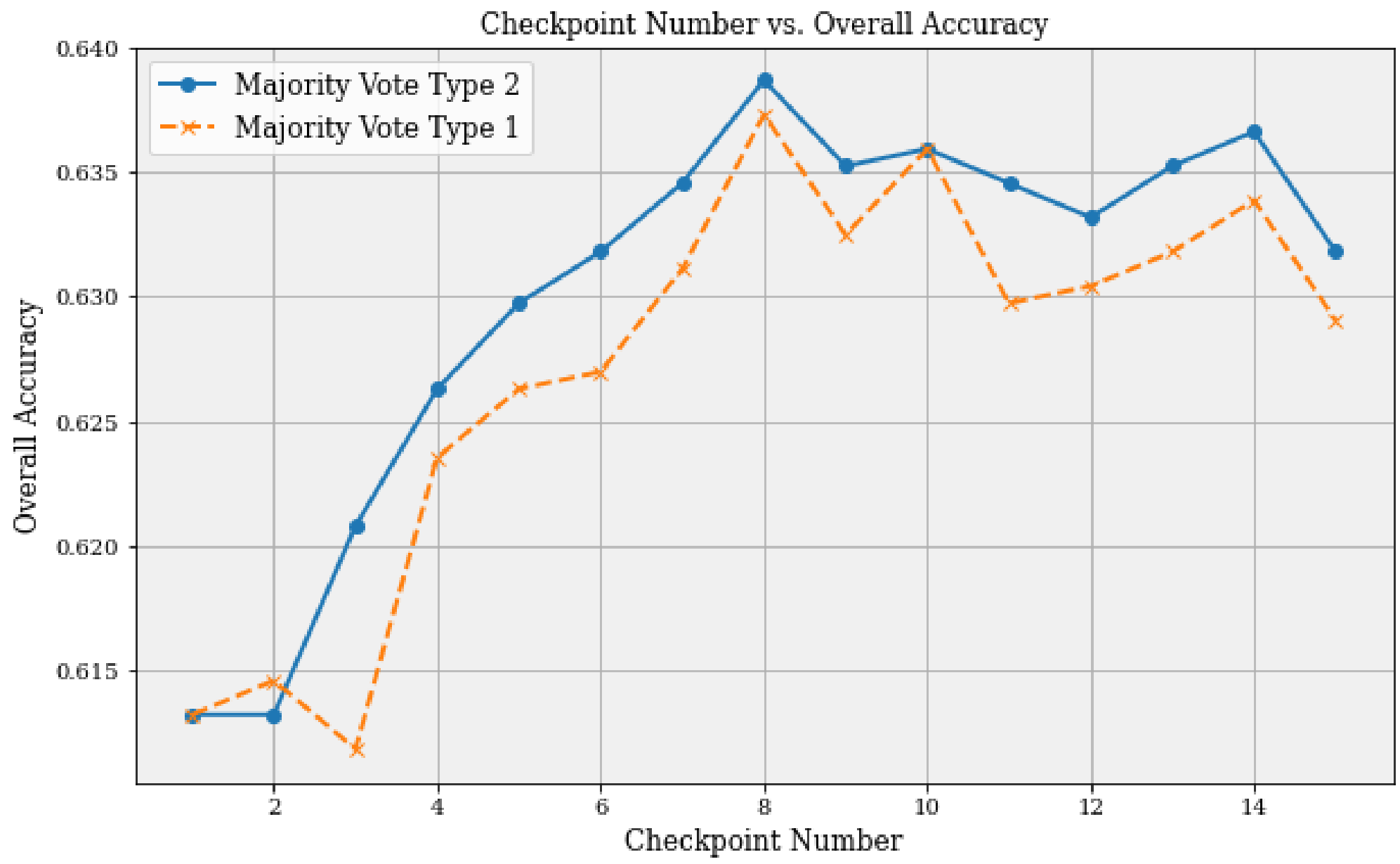

**Figure 7. Relationships between the number of checkpoints considered in the majority vote and the overall accuracy**

### 4.4 Impacts of prompt

The prompt is a key interface for users to communicate with LLMs, and designing effective prompts can significantly influence the performance and utility of these models (Liu et al., 2023). We further evaluate the impacts of the prompt on the performance of fine-tuned LLMs. The prompt template introduced in Llama fine-tuned Chat Models project (https://github.com/meta-llama/llama) was further considered (See prompt template type 5 in Appendix for details).

The top 5 checkpoints, which are fine-tuned using prompt templates from type 1 to type 4, are selected to evaluate the impacts of using different prompt templates (template type 5 was used) in the inference stage. Similar to previous experiments, the response of each test sample is regenerated until there is no "None" value in the corresponding response or it reaches five regeneration times. If the template used in the inference stage is different from that used in the fine-tuning stage, the generated response will not obey the human settings (Fig.8), and a significant decrease in model performance with the highest decrease ratio of more than 50%. The significant decrease in model performance emphasizes that the consistency between the fine-tuning prompts and inference prompts is crucial for optimal performance, indicating the fine-tuned LLM's reliance on specific prompt structures (how the task is presented) that it was trained on and it is highly sensitive to the input prompt structure. This shows us that the model learns not just the task but also the specific way in which the task is presented.

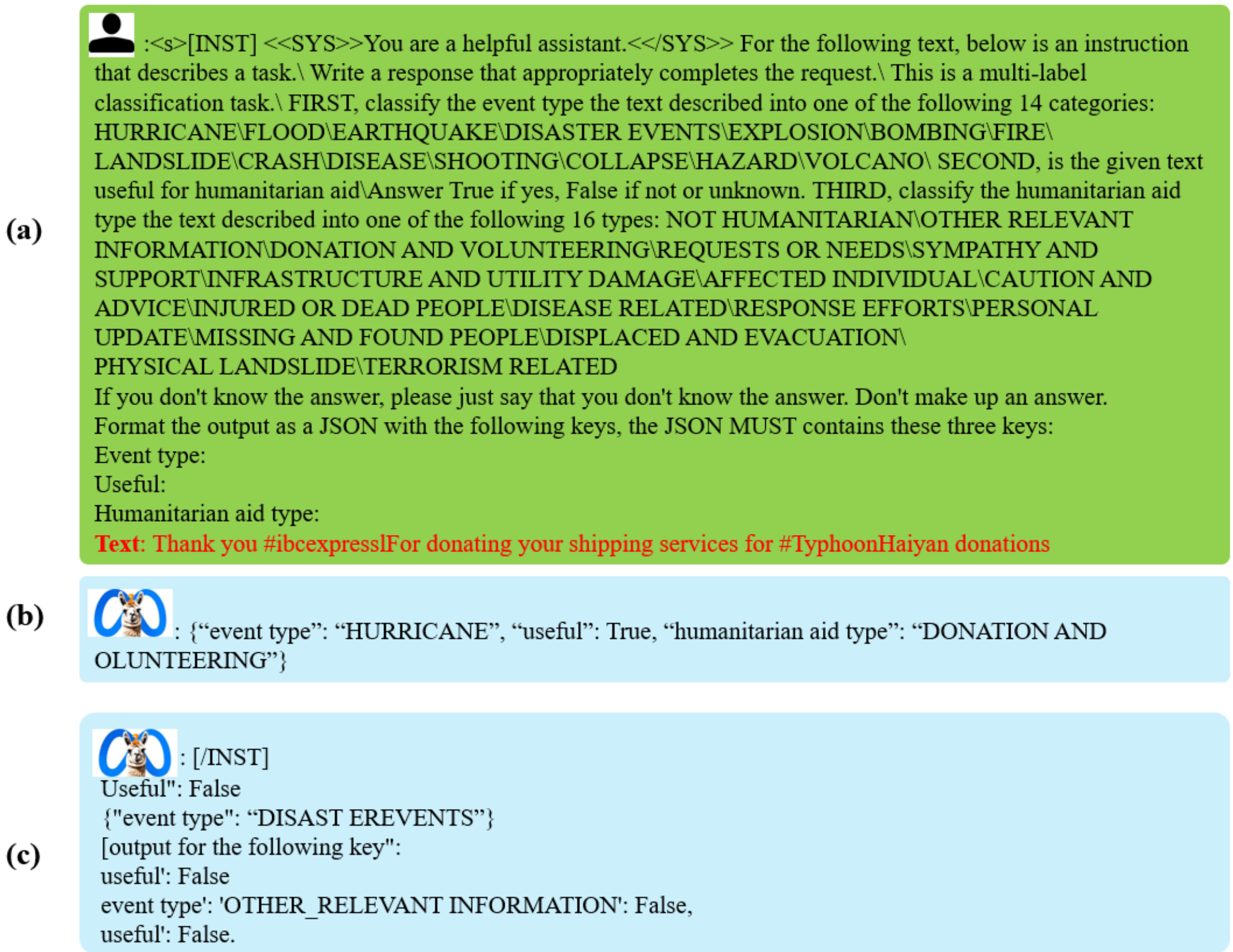

**Figure 8. Comparisons of response when generated using different and the same prompt template in the inference stage as in the fine-tuning stage.** (a) is the input for the fine-tuned model checkpoints; (b), (c) is the response generated using the same and different prompt template in the inference stage as in the fine-tuning stage.

**Table 2 The comparison of model overall accuracy between using the same and different prompt templates in inference stage as in the fine-tuning stage.** The model name follows the format of fine tuning strategy_(Lora_rank)_epochs.

| Model Name | Same template | Different template | Decrease ratio |
|---|---|---|---|
| Full_tune_0.315 | 0.613 | 0.524 | 14.5% |
| Chat_Lora_32_1 | 0.593 | 0.280 | 52.8% |
| Full_tune_0.265 | 0.585 | 0.427 | 27.0% |
| Full_tune_0.15 | 0.584 | 0.438 | 25.0% |
| Full_tune_0.135 | 0.581 | 0.345 | 40.6% |

## 5 Concluding Remarks

Automated extraction of situational information from social media is a crucial component of social sensing in disaster informatics. By overcoming the limitations of one-label classification methods, this study presents a significant advancement in the field of text classification in disaster informatics through harnessing multi-label classification capabilities of large language models (LLMs) through instruction fine-tuning. This research provides several key theoretical contributions to the field of disaster informatics and natural language processing. First, it introduces a novel methodology for fine-tuning pre-trained Large Language Models (LLMs) specifically for multi-label classification tasks in the context of disaster-related social media data. This approach addresses the limitations of traditional single-label classification methods, which often fail to capture the complex, multifaceted nature of social media posts during disasters. By enhancing the LLM with domain-specific knowledge, our study demonstrates the potential for more nuanced and comprehensive text classification, which is critical for effective disaster response and situational awareness. Second, this research contributes to the growing body of literature on the application of LLMs in specialized domains. While prior studies have explored the use of LLMs for various natural language processing tasks, our study is among the first to apply these models to the domain of disaster informatics. This opens new avenues for future research to explore the applicability of LLMs in other specialized fields, thereby broadening the scope of LLM applications and enhancing their utility across diverse contexts. Third, our research underscores the importance of multi-label classification in extracting actionable insights from social media data. The ability to simultaneously classify multiple aspects of a single post—such as the type of event, informativeness, and involvement of human aid—enhances the granularity and relevance of the information extracted. This multi-label approach not only improves the accuracy of situational assessments but also highlights the interconnectedness of different information types, which can provide more holistic insights during disaster events.

From a practical perspective, the findings of this research have significant implications for emergency management, public officials, and humanitarian organizations. The fine-tuned LLM developed in this study can be utilized as a powerful tool for real-time analysis of social media posts during disasters. By automating the classification of critical information, this model enables rapid situational awareness, which is essential for timely and effective response efforts. The practical benefits of this research are manifold:

1.Enhanced situational awareness: The fine-tuned model provides a more accurate and detailed understanding of disaster situations by categorizing social media posts into multiple relevant labels. This allows emergency responders to quickly identify and prioritize critical information, improving their ability to address the most urgent needs.

2.Resource efficiency: By reducing the redundancy and inefficiency associated with single-label models, our multi-label classification approach streamlines the data processing workflow. This efficiency gain is particularly valuable in disaster scenarios, where time and resources are often limited.

3. Scalability and adaptability: The methodology developed in this study can be adapted to other types of disasters and social media platforms, making it a versatile tool for various emergency management contexts. The scalability of our approach ensures that it can handle large volumes of data, which is crucial during widespread disaster events.

4.Policy and decision-making support: The insights generated by our model can inform policy decisions and strategic planning for disaster preparedness and response. By providing a data-driven basis for decision-making, our research supports more informed and effective policy interventions.

However, our research also uncovered several challenges, such as the sensitivity of the fine-tuned models to prompt structures and the risk of overfitting, which can limit their practical deployment. Also, the focus of this study was only on textual information within social media posts. Future studies can enhance the LLM presented in this study to also perform tasks on visual information available in social media posts to complement the textual information. Addressing these challenges requires further research to develop models that can handle variability in input data and operate effectively under different operational contexts. These advancements could foster more adaptive, resilient, and capable AI systems, ensuring that LLMs continue to serve as invaluable assets in enhancing situational awareness in disaster response. Our work lays a foundational step towards this future, offering the first fine-tuned LLM model for disaster informatics to inform subsequent efforts to harness AI in improving public safety in disasters.

# Appendix

## Prompt template—type1:

<s>### **Instruction**:
For the following text, below is an instruction that describes a task.\
Write a response that appropriately completes the request.\
INSTRUCTION: This is a classification task.\
Classify the event type the text described into one of the following 14 categories:
HURRICANE\nFLOOD\nEARTHQUAKE\nDISASTER\nEVENTS\nEXPLOSION\nBOMBIN G\nFIRE\nLANDSLIDE\nCRASH\nDISEASE\nSHOOTING\nCOLLAPSE\nHAZARD\nVOLC ANO\n
Format the output as JSON with the following keys:
event type
text: {text}
### **Response**:
{REPONSE} </s>

## Prompt template—type2:

<s>**### Instruction**:
For the following text, below is an instruction that describes a task.\
Write a response that appropriately completes the request.\
INSTRUCTION: This is a classification task.\
Is the given text useful for humanitarian aid\
Answer True if yes, False if not or unknown.\
Format the output as JSON with the following key:
useful
text: {text}
**### Response**:
{REPONSE} </s>

## Prompt template—type3:

<s>### Instruction:
For the following text, below is an instruction that describes a task.\
Write a response that appropriately completes the request.\
INSTRUCTION: This is a classification task.\
Classify the humanitarian aid type the text described into one of the following 16 types:
NOT HUMANITARIAN\nOTHER RELEVANT INFORMATION\nDONATION AND VOLUNTEERING\nREQUESTS OR NEEDS\nSYMPATHY AND SUPPORT\nINFRASTRUCTURE AND UTILITY DAMAGE\nAFFECTED INDIVIDUAL\nCAUTION AND ADVICE\nINJURED OR DEAD PEOPLE\nDISEASE RELATED\nRESPONSE EFFORTS\nPERSONAL UPDATE\nMISSING AND FOUND PEOPLE\nDISPLACED AND EVACUATION\nPHYSICAL LANDSLIDE\nTERRORISM RELATED
Format the output as JSON with the following keys:
humanitarian aid type
text: {text}
### Response:
{REPONSE} </s>

## Prompt template—type4:

<s>### **Instruction**:
For the following text, below is an instruction that describes a task.\
Write a response that appropriately completes the request.\
This is a multi-label classification task.\
FIRST, classify the event type the text described into one of the following 14 categories:
HURRICANE\nFLOOD\nEARTHQUAKE\nDISASTER\EVENTS\nEXPLOSION\nBOMBING\ nFIRE\LANDSLIDE\nCRASH\nDISEASE\nSHOOTING\nCOLLAPSE\nHAZARD\nVOLCAN

O\n

SECOND, is the given text useful for humanitarian aid\
Answer True if yes, False if not or unknown.\n

THIRD, classify the humanitarian aid type the text described into one of the following 16 types:
NOT HUMANITARIAN\nOTHER RELEVANT INFORMATION\nDONATION AND VOLUNTEERING\nREQUESTS OR NEEDS\nSYMPATHY AND SUPPORT\INFRASTRUCTURE AND UTILITY DAMAGE\nAFFECTED\n INDIVIDUAL\nCAUTION AND ADVICE\nINJURED OR DEAD PEOPLE\
DISEASE RELATED\nRESPONSE EFFORTS\nPERSONAL UPDATE\nMISSING AND FOUND PEOPLE\nDISPLACED AND EVACUATION\PHYSICAL LANDSLIDE\nTERRORISM RELATED
If you don't know the answer, please just say that you don't know the answer. Don't make up an answer.
Format the output as a JSON with the following keys, the JSON MUST contains these three keys:
Event type:
Useful:
Humanitarian aid type:
text: {text}
Response:
{REPONSE} </s>

### Prompt template—type5:

<s>[INST] <<SYS>>
You are a helpful assistant.
<</SYS>>
For the following text, below is an instruction that describes a task.\
Write a response that appropriately completes the request.\
This is a multi-label classification task.\
FIRST, classify the event type the text described into one of the following 14 categories:
HURRICANE\nFLOOD\nEARTHQUAKE\nDISASTER EVENTS\nEXPLOSION\nBOMBING\nFIRE\
LANDSLIDE\nCRASH\nDISEASE\nSHOOTING\nCOLLAPSE\nHAZARD\nVOLCANO\
SECOND, is the given text useful for humanitarian aid\
Answer True if yes, False if not or unknown.\n
THIRD, classify the humanitarian aid type the text described into one of the following 16 types:
NOT HUMANITARIAN\nOTHER RELEVANT INFORMATION\nDONATION AND VOLUNTEERING\nREQUESTS OR NEEDS\nSYMPATHY AND SUPPORT\
INFRASTRUCTURE AND UTILITY DAMAGE\nAFFECTED INDIVIDUAL\nCAUTION AND ADVICE\nINJURED OR DEAD PEOPLE\
DISEASE RELATED\nRESPONSE EFFORTS\nPERSONAL UPDATE\nMISSING AND FOUND PEOPLE\nDISPLACED AND EVACUATION\
PHYSICAL LANDSLIDE\nTERRORISM RELATED

If you don't know the answer, please just say that you don't know the answer. Don't make up an answer.
Format the output as a JSON with the following keys, the JSON MUST contains these three keys:
Event type:
Useful:
Humanitarian aid type:

Text: {text}
Response:
[/INST]
{REPONSE} </s>

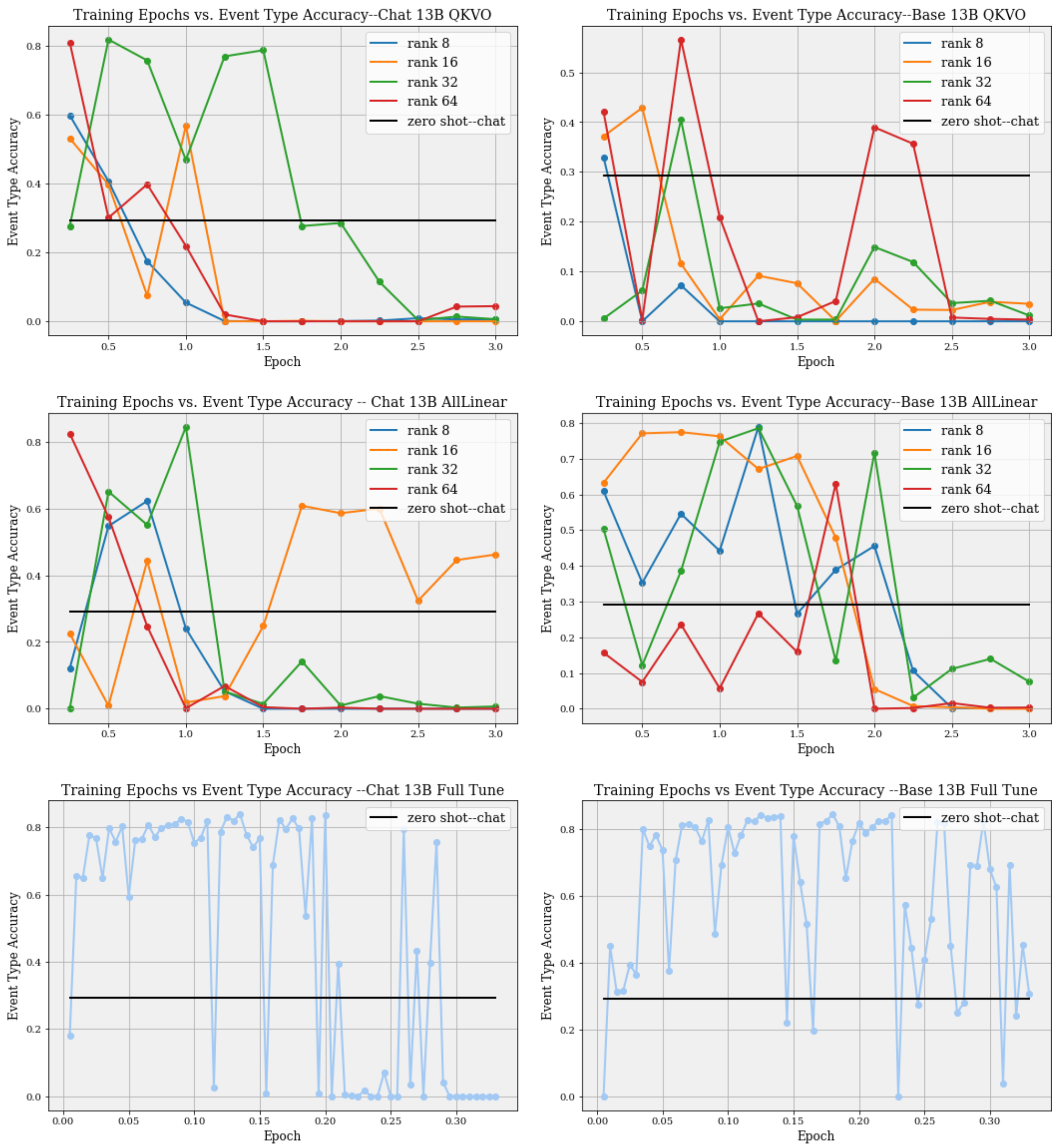

**Figure 9. The event type accuracy of the LORA fine tuning and full parameter tuning.** The event type accuracy of zero-shot is from chat foundation model.

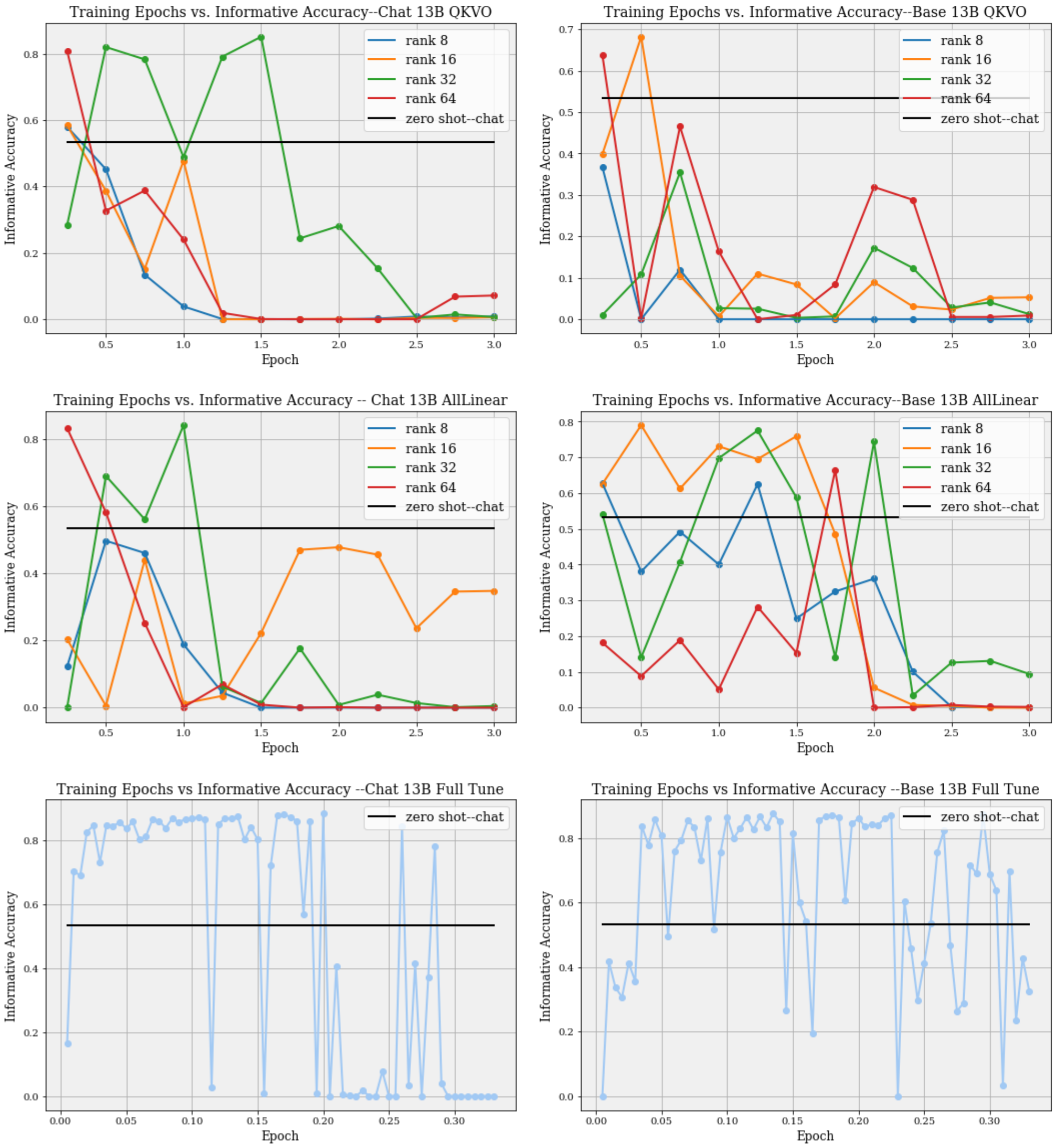

**Figure 10. The informative accuracy of the LORA fine tuning and full parameter tuning.**
The informative accuracy of zero-shot is from chat foundation model.

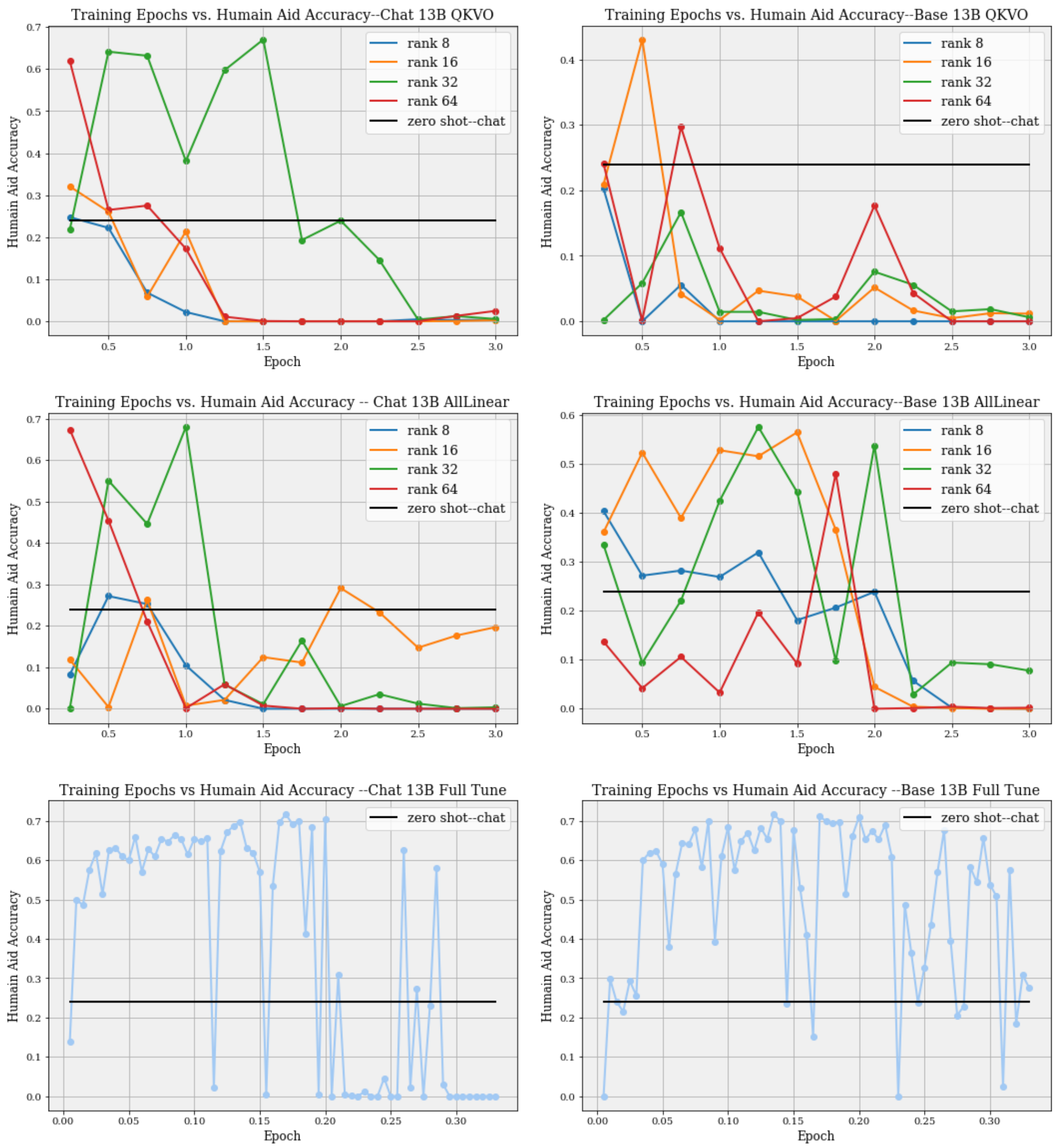

**Figure 11. The human aid type accuracy of the LORA fine tuning and full parameter tuning.** The human aid type accuracy of zero-shot is from chat foundation model.

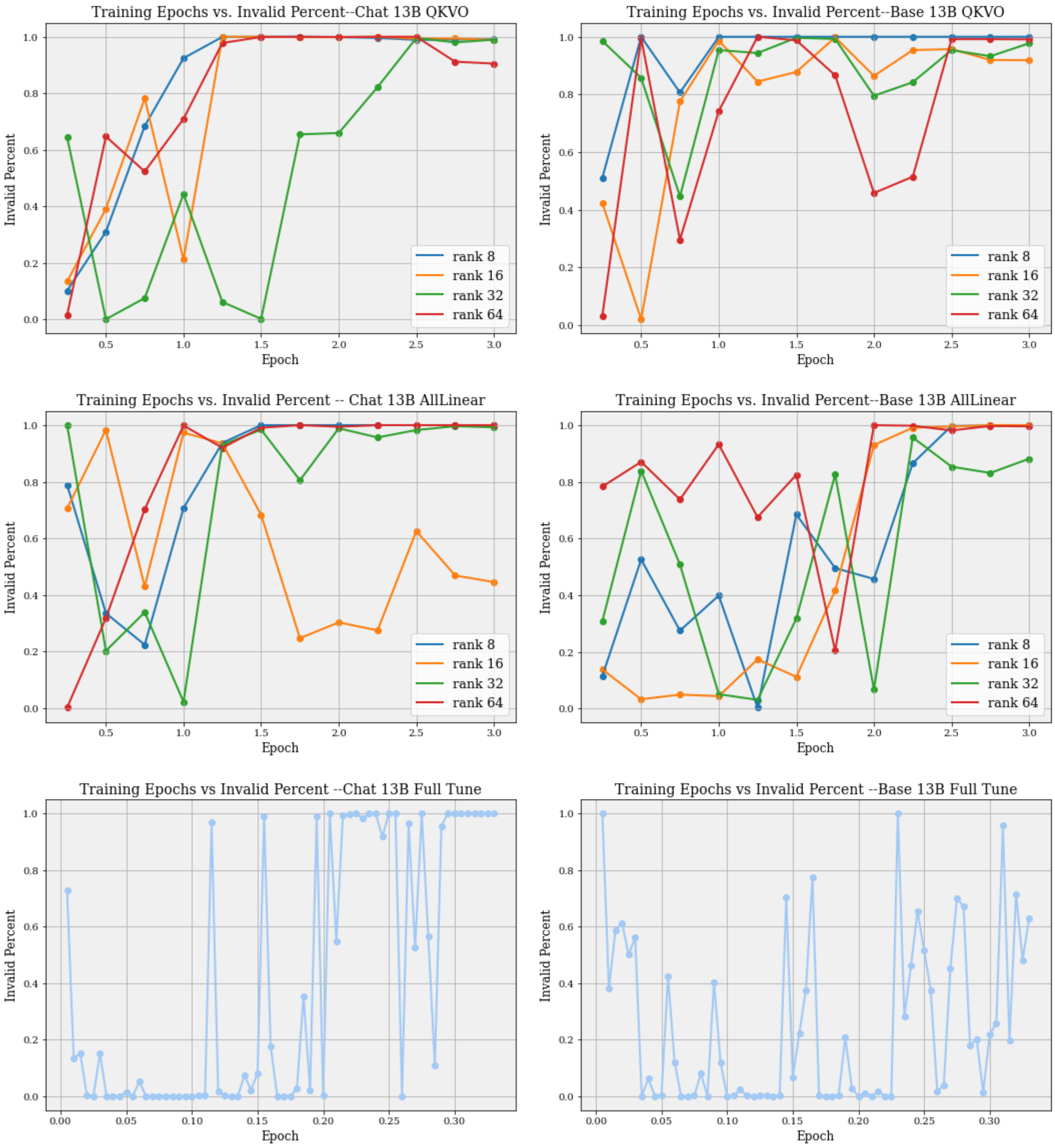

**Figure 12. The percent of invalid response of the LORA fine tuning and full parameter tuning.**

**Data Availability**
The dataset used in this paper are publicly accessible and cited in this paper.

**Code Availability**
The code that supports the findings of this study is available from the corresponding author upon request.

**Acknowledgements**
This work was supported by National Science Foundation under CRISP 2.0 Type 2 No. 1832662 grant. Any opinions, findings, conclusions, or recommendations expressed in this research are those of the authors and do not necessarily reflect the view of the funding agencies. Portions of this research were conducted with the advanced computing resources provided by Texas A&M High Performance Research Computing. We would like to thank Guanchu Wang and Hongyi Liu from Department of Computer Science in Rice University for their scientific and vital suggestions on this work.

**Author contributions**
**Kai Yin**: Conceptualization, Methodology, Software, Formal analysis, Investigation; Writing, original draft; Visualization; **Chengkai Liu:** Conceptualization, Methodology, Formal analysis, Investigation; Writing, original draft; **Ali Mostafavi**: Conceptualization; Methodology; Writing, Reviewing and Editing; Supervision; Funding acquisition; **Xia Hu:** Conceptualization; Methodology; Writing, Reviewing and Editing; Supervision

**Competing interests**

The authors declare no competing interests.

**Additional information**
Supplementary material associated with this article can be found in the attached document.

## Reference

Alam, F., Ofli, F., & Imran, M. (2016). *CrisisMMD : Multimodal Twitter Datasets from Natural Disasters*.

Alam, F., Sajjad, H., Imran, M., & Ofli, F. (2021). *CrisisBench: Benchmarking Crisis-related Social Media Datasets for Humanitarian Information Processing*.

Brown, T. B., Kaplan, J., Ryder, N., Henighan, T., Chen, M., Herbert-voss, A., Ziegler, D. M., Krueger, G., Askell, A., Hesse, C., & Mccandlish, S. (2020). *Language Models are Few-Shot Learners*.

Collins, C., Dennehy, D., Conboy, K., & Mikalef, P. (2021). Artificial intelligence in information systems research: A systematic literature review and research agenda. *International Journal of Information Management*, *60*(June), 102383. https://doi.org/10.1016/j.ijinfomgt.2021.102383

Dong, L., Yang, N., Wang, W., Wei, F., Liu, X., & Wang, Y. (2019). *Unified Language Model Pre-training for Natural Language Understanding and Generation*. *NeurIPS*.

Dubey, R., Bryde, D. J., Dwivedi, Y. K., Graham, G., & Foropon, C. (2022). Impact of artificial intelligence-driven big data analytics culture on agility and resilience in humanitarian supply chain: A practice-based view. *International Journal of Production Economics*, *250*(March), 108618. https://doi.org/10.1016/j.ijpe.2022.108618

Fan, C., & Mostafavi, A. (2020). A graph-based method for social sensing of infrastructure disruptions in disasters. *Computers, Environment and Urban Systems*.

Fan, C., Wu, F., & Mostafavi, A. (2020). A Hybrid Machine Learning Pipeline for Automated Mapping of Events and Locations from Social Media in Disasters. *IEEE Access*, *8*, 10478–10490. https://doi.org/10.1109/ACCESS.2020.2965550

Fan, C., Zhang, C., Yahja, A., & Mostafavi, A. (2021). Disaster City Digital Twin: A vision for integrating artificial and human intelligence for disaster management. *International Journal of Information Management*, *56*(March 2019), 102049. https://doi.org/10.1016/j.ijinfomgt.2019.102049

Fu, J., Han, H., Su, X., & Fan, C. (2023). *Towards Human-AI Collaborative Urban Science Research Enabled by Pre-trained Large Language Models*. 1–11.

Gao, Y., Xiong, Y., Gao, X., Jia, K., Pan, J., Bi, Y., Dai, Y., & Sun, J. (2024). *Retrieval-Augmented Generation for Large Language Models : A Survey*. 1–21.

Grootendorst, M. (2023). *3 Easy Methods For Improving Your Large Language Model*. https://towardsdatascience.com/3-easy-methods-for-improving-your-large-language-model-68670fde9ffa

Gupta, S., Chen, Y., & Tsai, C.-H. (2024). *Utilizing Large Language Models in Tribal Emergency Management*. 1–6. https://doi.org/10.1145/3640544.3645219

Han, Z., Gao, C., Liu, J., Zhang, J. J., & Zhang, S. Q. (2024). *Parameter-Efficient Fine-Tuning for Large Models : A Comprehensive Survey*. 1–24.

Hu, E., Shen, Y., Wallis, P., Allen-Zhu, Z., Li, Y., & Wang, S. (2021). *LORA: LOW-RANK ADAPTATION OF LARGE LANGUAGE MODELS*. 1–26.

Kaplan, J., Wu, J., Amodei, D., Henighan, T., Gray, S., Brown, T. B., & Radford, A. (2020). *Scaling Laws for Neural Language Models*.

Lee, C. C., Maron, M., & Mostafavi, A. (2022). Community-scale big data reveals disparate impacts of the Texas winter storm of 2021 and its managed power outage. *Humanities and Social Sciences Communications*, *9*(1). https://doi.org/10.1057/s41599-022-01353-8

Lester, B., Al-Rfou, R., & Constant, N. (2021). The Power of Scale for Parameter-Efficient Prompt Tuning. *EMNLP 2021 - 2021 Conference on Empirical Methods in Natural Language Processing, Proceedings*, 3045–3059. https://doi.org/10.18653/v1/2021.emnlp-main.243

Li, X. L., & Liang, P. (2021). Prefix-tuning: Optimizing continuous prompts for generation. *ACL-IJCNLP 2021 - 59th Annual Meeting of the Association for Computational Linguistics and the 11th International Joint Conference on Natural Language Processing, Proceedings of the Conference*, 4582–4597. https://doi.org/10.18653/v1/2021.acl-long.353

Liang, Y., Liu, Y., Wang, X., & Zhao, Z. (2023). *Exploring Large Language Models for Human Mobility Prediction under Public Events*.

Liu, X., Wang, J., Sun, J., Yuan, X., Dong, G., Di, P., Wang, W., & Wang, D. (2023). Prompting Frameworks for Large Language Models : A Survey. *ACM Computing Surveys*, *1*(1), 1–34.

Liu, Z., Yin, K., & Mostafavi, A. (2024). Rethinking Urban Flood Risk Assessment By Adapting Health Domain Perspective. *ArXiv:2403.03996*.

Liu, Z., Zhang, A., Yao, Y., Shi, W., Huang, X., & Shen, X. (2021). Analysis of the performance and robustness of methods to detect base locations of individuals with geo-tagged social media data.

*International Journal of Geographical Information Science*, *35*(3), 609–627. https://doi.org/10.1080/13658816.2020.1847288
Liu, Z., Zhou, X., Shi, W., & Zhang, A. (2018). Towards detecting social events by mining geographical patterns with VGI data. *ISPRS International Journal of Geo-Information*, *7*(12). https://doi.org/10.3390/ijgi7120481
Luo, Y., Yang, Z., Meng, F., Li, Y., Zhou, J., & Zhang, Y. (2024). *An Empirical Study of Catastrophic Forgetting in Large Language Models During Continual Fine-tuning*. 1–14.
Quentin, M. (2019). *Parameter-Efficient Transfer Learning for NLP*.
Scao, T. Le, Fan, A., Akiki, C., Pavlick, E., Ili, S., Hesslow, D., Luccioni, A. S., Gallé, M., Tow, J., Alexander, M., Biderman, S., Webson, A., Ammanamanchi, P. S., Wang, T., Muennighoff, N., Villanova, A., Ruwase, O., Bawden, R., Mcmillan-major, A., … Castagné, R. (2023). *BLOOM: A 176B-Parameter Open-Access Multilingual Language Model*.
Schuurmans, D. (2023). *LEAST-TO-MOST PROMPTING ENABLES COMPLEX REASONING IN LARGE LANGUAGE MODELS*. 1–61.
Shi, W., Liu, Z., An, Z., & Chen, P. (2021). RegNet: a neural network model for predicting regional desirability with VGI data. *International Journal of Geographical Information Science*, *35*(1), 175–192. https://doi.org/10.1080/13658816.2020.1768261
Shoeybi, M., Patwary, M., Puri, R., Legresley, P., Casper, J., & Catanzaro, B. (2018). *Megatron-LM: Training Multi-Billion Parameter Language Models Using Model Parallelism*. https://github.com/
Sun, X., Li, X., Li, J., Wu, F., Guo, S., Zhang, T., & Wang, G. (2023). *Text Classification via Large Language Models*. http://arxiv.org/abs/2305.08377
Suwaileh, R., Elsayed, T., & Imran, M. (2023). IDRISI-RE: A generalizable dataset with benchmarks for location mention recognition on disaster tweets. *Information Processing and Management*, *60*(3), 103340. https://doi.org/10.1016/j.ipm.2023.103340
Suwaileh, R., Elsayed, T., Imran, M., & Sajjad, H. (2022). When a disaster happens, we are ready: Location mention recognition from crisis tweets. *International Journal of Disaster Risk Reduction*, *78*(August 2021), 103107. https://doi.org/10.1016/j.ijdrr.2022.103107
Taori, R., Gulrajani, I., Zhang, T., Dubois, Y., Li, X., Guestrin, C., Liang, P., & Hashimoto, T. B. (2023). Stanford Alpaca : An Instruction-following LLaMA Model. *GitHub Repository*, 1–7. https://github.com/tatsu-lab/stanford_alpaca
Tay, Y., Wei, J., Zheng, H. S., Garcia, X., & Zhou, D. (2022). *Transcending Scaling Laws with 0.1% Extra Compute*.
Touvron, H., Lavril, T., Izacard, G., Martinet, X., Lachaux, M.-A., Lacroix, T., Rozière, B., Goyal, N., Hambro, E., Azhar, F., Rodriguez, A., Joulin, A., Grave, E., & Lample, G. (2023). *LLaMA: Open and Efficient Foundation Language Models*. http://arxiv.org/abs/2302.13971
Touvron, H., Martin, L., Stone, K., Albert, P., Almahairi, A., Babaei, Y., Bashlykov, N., Batra, S., Bhargava, P., Bhosale, S., Bikel, D., Blecher, L., Ferrer, C. C., Chen, M., Cucurull, G., Esiobu, D., Fernandes, J., Fu, J., Fu, W., … Scialom, T. (2023). *Llama 2: Open Foundation and Fine-Tuned Chat Models*. http://arxiv.org/abs/2307.09288
Vaswani, A., Shazeer, N., Parmar, N., Uszkoreit, J., Jones, L., Gomez, A. N., Kaiser, Ł., & Polosukhin, I. (2017). Attention is all you need. *Advances in Neural Information Processing Systems*, *2017-Decem*(Nips), 5999–6009.
Wang, T., & Hesslow, D. (2022). *What Language Model Architecture and Pretraining Objective Work Best for Zero-Shot Generalization ?* 1–26.
Wang, W., Yang, S., Yin, K., Zhao, Z., Ying, N., & Fan, J. (2022). Network approach reveals the spatiotemporal influence of traffic on air pollution under COVID-19. *Chaos: An Interdisciplinary Journal of Nonlinear Science*, *January 2019*. https://doi.org/10.1063/5.0087844
Wang, Y., & Yu, P. S. (2023). *Named Entity Recognition via Machine Reading Comprehension: A Multi-Task Learning Approach*.
Wei, J., Schuurmans, D., Chi, E. H., Le, Q. V, & Zhou, D. (2022). *Chain-of-Thought Prompting Elicits Reasoning in Large Language Models*. *NeurIPS*, 1–14.
Ye, S., Hwang, H., Yang, S., Yun, H., Kim, Y., & Seo, M. (2023). *In-Context Instruction Learning*. http://arxiv.org/abs/2302.14691
Yin, K., & Mostafavi, A. (2023a). Deep Learning-driven Community Resilience Rating based on Intertwined Socio-Technical Systems Features. *ArXiv:2311.01661*.
Yin, K., & Mostafavi, A. (2023b). Unsupervised Graph Deep Learning Reveals Emergent Flood Risk Profile of Urban Areas. *Aeprint ArXiv:2309.14610*. https://doi.org/10.48550/arXiv.2309.14610
Yin, K., Wu, J., Wang, W., Lee, D., & Wei, Y. (2023). An integrated resilience assessment model of urban transportation network : A case study of 40 cities in China. *Transportation Research Part*

*A*, *173*(March 2021), 103687. https://doi.org/10.1016/j.tra.2023.103687
Zeng, A., Liu, X., Du, Z., Wang, Z., & Lai, H. (2023). *GLM-130B: AN OPEN BILINGUAL PRE-TRAINED MODEL*. 1–56.
Zhang, B., Ghorbani, B., Bapna, A., Cheng, Y., & Garcia, X. (2020). *Examining Scaling and Transfer of Language Model Architectures*.
Zhang, C., Fan, C., Yao, W., Hu, X., & Mostafavi, A. (2019). Social media for intelligent public information and warning in disasters: An interdisciplinary review. *International Journal of Information Management*, *49*(April), 190–207. https://doi.org/10.1016/j.ijinfomgt.2019.04.004
Zhang, Shengyu, Dong, L., Li, X., Zhang, S., Sun, X., Wang, S., Li, J., Hu, R., Zhang, T., Wu, F., & Wang, G. (2023). *Instruction Tuning for Large Language Models: A Survey*. *2*. http://arxiv.org/abs/2308.10792
Zhang, Susan, Roller, S., Artetxe, M., Chen, M., Chen, S., Dewan, C., Diab, M., Li, X., Lin, X. V., Mihaylov, T., Ott, M., Shleifer, S., Shuster, K., Simig, D., Koura, P. S., Sridhar, A., Wang, T., & Zettlemoyer, L. (2023). *OPT : Open Pre-trained Transformer Language Models*.
Zhang, X., & Zhang, Q. (2021). *Enhancing Label Correlation Feedback in Multi-Label Text Classification via Multi-Task Learning*.
Zhao, W. X., Zhou, K., Li, J., Tang, T., Wang, X., Hou, Y., Min, Y., Zhang, J., Dong, Z., Du, Y., Yang, C., Chen, Y., Chen, Z., Jiang, J., Ren, R., Li, Y., Tang, X., Liu, Z., Liu, P., … Wen, J. (2023). *A Survey of Large Language Models*. 1–124.